\documentclass[letterpaper, 10 pt, conference]{ieeeconf}  

\IEEEoverridecommandlockouts                              

\usepackage{graphicx} 
\usepackage{amsmath} 

\usepackage{amsthm}
\usepackage{listings}
\theoremstyle{definition}

\usepackage{amssymb}
\usepackage{pifont}
\usepackage{booktabs}

\usepackage{algorithm}
\usepackage[noend]{algpseudocode}
\usepackage[colorlinks=true]{hyperref}
\usepackage[font=scriptsize,labelfont=bf]{caption}
\usepackage{etoolbox}
\makeatletter
\patchcmd{\@makecaption}
  {\scshape}
  {}
  {}
  {}
\makeatother

\title{\LARGE \bf
Automated Generation of Robotic Planning Domains from Observations 
}

\author{Maximilian Diehl$^{1}$, Chris Paxton$^{2}$, and Karinne Ramirez-Amaro$^{1}$ 
\thanks{$^{1}$Maximilian Diehl and Karinne Ramirez-Amaro. Faculty of Electrical Engineering, Chalmers University of Technology, SE-412 96 Gothenburg, Sweden.
        {\tt\small \{diehlm, karinne\}@chalmers.se}}%
\thanks{$^{2}$Chris Paxton is with NVIDIA, USA.
        {\tt\small cpaxton@nvidia.com}}%
}

\begin{document}

\maketitle
\thispagestyle{empty}
\pagestyle{empty}

\begin{abstract} 
Automated planning enables robots to find plans to achieve complex, long-horizon tasks, given a planning domain. This planning domain consists of a list of actions, with their associated preconditions and effects, and is usually manually defined by a human expert, which is very time-consuming or even infeasible. In this paper, we introduce a novel method for generating this domain automatically from human demonstrations. First, we automatically segment and recognize the different observed actions from human demonstrations. From these demonstrations, the relevant preconditions and effects are obtained, and the associated planning operators are generated. Finally, a sequence of actions that satisfies a user-defined goal can be planned using a symbolic planner. The generated plan is executed in a simulated environment by the TIAGo robot. We tested our method on a dataset of 12 demonstrations collected from three different participants. The results show that our method is able to generate executable plans from using one single demonstration with a 92\% success rate, and 100\% when the information from all demonstrations are included, even for previously unseen stacking goals.
\end{abstract}

\section{Introduction}
\addcontentsline{toc}{section}{Introduction}
\label{sec:intro}
One of the most important competencies of autonomous robots is the ability to learn from experience~\cite{ADL15}, refine and combine skills~\cite{ADL10} to reuse knowledge for unseen tasks. From the beginnings of AI, Automated Planning (AP) has been playing an important role in achieving this deliberation of autonomous robots~\cite{INGRAND201710}. AP is the process of planning ahead on how to reach a goal based on a set of available and applicable high-level actions~\cite{ADL3}. The abstraction from lower-level activities into high-level actions is advantageous for complex and long-horizon planning tasks~\cite{ADL9}, even in continuous real-world-domains. 

Several platforms like the Sequence-Planner~\cite{sp} or ROSPlan~\cite{ROSPlan} provide sophisticated planning and plan execution capabilities. However, one of the major bottlenecks is the requirement of an accurate description of the planning task, also called the planning domain. The planning domain involves a list of all possible actions/operators defined via name, a set of preconditions, and a set of effects. Generating large domains by hand is very time-consuming or even infeasible~\cite{jimenez2012review}.

\begin{figure}[ht!]
\centering
  \includegraphics[width=0.48\textwidth]{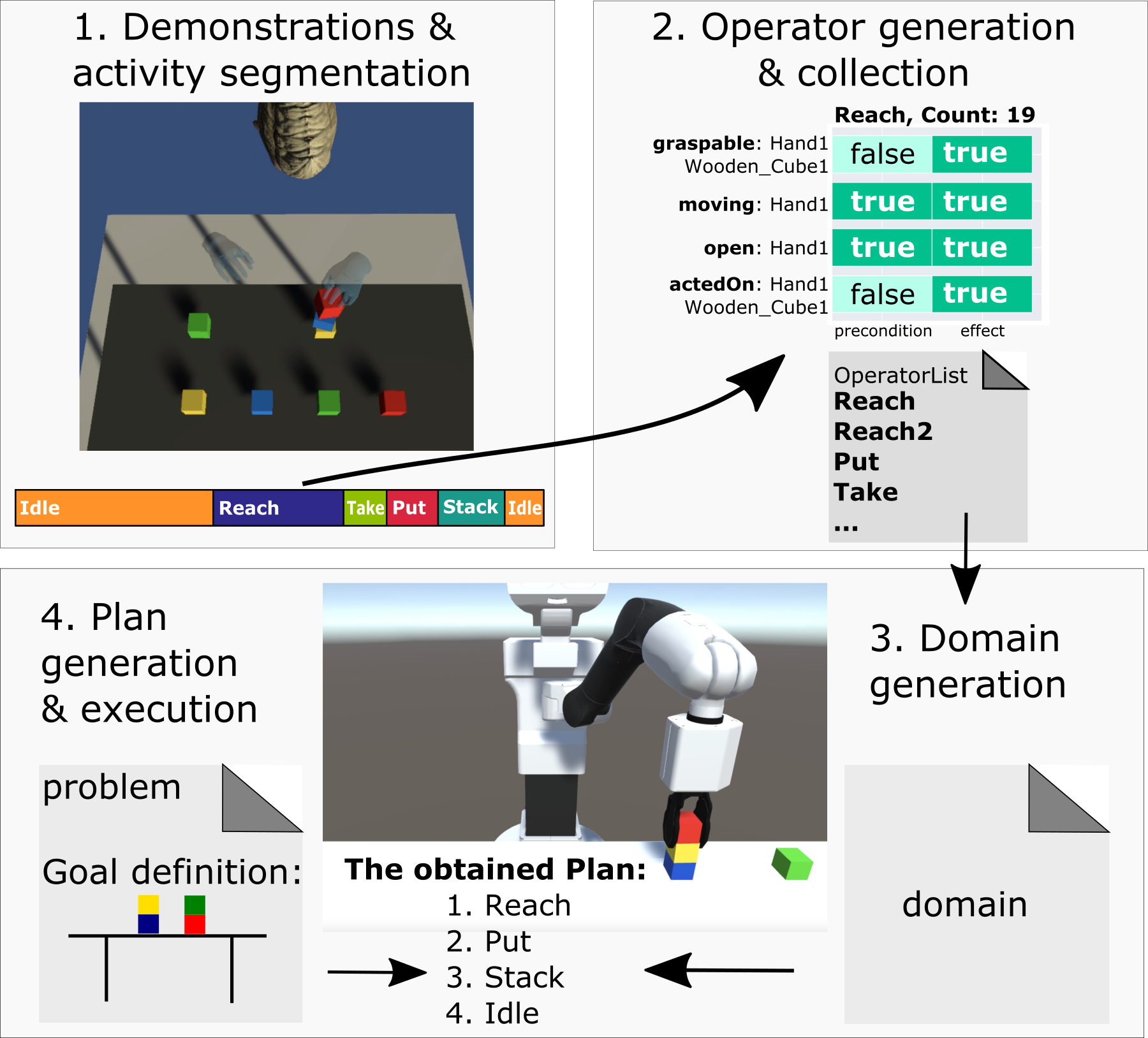}
  \caption{This figure illustrates the individual blocks of our system. 1. The domain actions are obtained from human demonstrations performed in Virtual Reality. 2. For the operator generation, actions are classified, and their preconditions and effects are extracted. 3. At run-time, a planning domain is generated from the operator list. 4. Given a user-defined goal state, the plan is constructed by an off-the-shelf symbolic planner like Fast-Downward and executed by the TIAGo robot.}
\label{fig:overview}
\end{figure}
We, therefore, propose a system for automated domain generation that builds on prior work on activity segmentation and classification from human demonstration in a Virtual Reality (VR) environment~\cite{BatesRIC17}. Our main contribution is the integration of this action recognition into a novel operator generation process that analyzes the transitions of the classified activities, extracts the relevant preconditions and effects, and converts the frequency of the demonstrated operators into an optimization criterion for the plan generation. As a proof-of-concept, we also implemented a PDDL~\cite{pddl} parser that enables plan creation with one of the many available off-the-shelf planners~\cite{fastdownward}. Finally, the obtained plans are executed in a simulated environment through the TIAGo robot (see Fig. \ref{fig:overview}).

There are several challenges that need to be addressed in order to generate a useful planning domain.
First, the action demonstrations can be noisy or incomplete. For this reason, we introduce an operator cost that prioritizes more commonly observed operators during plan generation. 
Additionally, our environment contains many objects and relations, which are irrelevant for a specific activity. For example, when picking up a specific \texttt{Cube\_green}, the other cubes
have no bearing on the execution of the action,
which should reflect into the predicate selection for preconditions and effects of the generated operator.
Lastly, operators should generalize activities performed with a specific hand or objects (e.g.  \texttt{Right\_hand},  \texttt{Cube\_green}) to types (e.g. \textit{Hand}, \textit{Wooden\_cube}).

Another advantage of our system is that the collection of operators can continuously grow with new demonstrations. 
As a result, the planning system was able to create plans for unseen goals in our experiments. While demonstrations 
only covered the stacking of one or two cubes in a row, our system was able to create plans for more complex tasks such as stacking four cubes or two separate towers.

To summarize, our main contributions are:
\begin{itemize}
    \item High-level operator generation from noisy demonstrations, including the omission of irrelevant preconditions/effects and generalization from objects to classes.
    \item Integration of the operator generation process with activity recognition from human demonstrations, plan generation, and execution procedure.
    \item Operator collection, which is automatically extended with each new demonstration, and prioritizes more often observed operators during the plan generation.
\end{itemize}

\section{Related Work}
\addcontentsline{toc}{section}{Related Work}
From a planning community perspective, the field of Knowledge Engineering is concerned with the acquisition and formulation of planning knowledge, with the domain model being the desired output \cite{ADL3}. There are various algorithms, like LOCM2\cite{locm}, or AMAN \cite{aman}, which try to generate operators, usually based on plan traces. Some methods also require a partial domain model, e.g., descriptions of objects, attributes, relations, and operator names. Plan traces, for example, in the form of random action sequences \cite{ADL11}, are usually constructed from benchmark domains, like the ones defined by the International Planning Competition (IPC). These action sequences are then used as input for the operator generation process. The quality of the newly constructed domain can be evaluated by comparing it back with the original domain used to generate the inputs. However, obtaining these traces from real-world demonstrations, an essential part of our system, is often not studied.

There are also some works from the robotics community that aim to learn planning operators. Many approaches do this directly on their respective robotic platforms. In~\cite{Ahmetoglu2020} a framework is presented that learns probabilistic action effects. As input, the framework requires the motor commands associated, for example, with the stacking activity, and possible outcomes are simulated in an environment with different objects like cubes and spheres. In another approach~\cite{ADL9} state representations and operators are learned while the robot is exploring its environment. In~\cite{ADL16}, and~\cite{ADL12} the technique of kinesthetic teaching is used to generate operators for skills like reaching, grasping, pushing, and pulling. Recent work also deals with the question of how to abstract the essence behind lower-level actions \cite{ADL8, ADL9, ADL10}. In~\cite{ADL10} a parameterized model of a pushing activity is learned, which relates the applied force to the final block location after sliding. The authors of \cite{ADL8} abstract tasks from the sensorimotor level to state variables like {\tt inFrontOf}, {\tt behind}, or {\tt above}. The task of opening a drawer and a door is learned based on these spatial and temporal features. The biggest difference is that our method learns from human demonstrations. Additionally, by utilizing our activity recognition framework~\cite{BatesRIC17} we can derive semantic operator descriptions and, most importantly, meaningful segmentation of the observed task, rather than considering every state change as a potential new operator.

Regarding plan generation and execution platforms, different alternatives are proposed, e.g., Sequence-Planner (SP)~\cite{sp}, ROSPlan~\cite{ROSPlan}, and CRAM~\cite{cram}.
While the SP has a direct interface to an SAT solver (propositional satisfiability problem-solving algorithm) and ROSPlan is based on the PDDL formulation, the third approach can be considered as a goal reasoning system, where under-specified plan executions are filled up with information from a knowledge base. Nevertheless, domain information must be manually supplied in all three of them. 

\section{Planning problem and environment}
\addcontentsline{toc}{section}{Planning problem and environment}
\subsection{Environment}
In this paper, we study the task of stacking cubes, which is a well-known problem in the planning community \cite{AI}. We draw a clear distinction between the demonstration and execution environments (both visualized in Fig. \ref{fig:overview}). Even though both environments consist of a table with cubes, we deal with different instances of objects. To cope with these differences, we introduce an Ontology that can describe different cube and table instances through general class names such as \texttt{Wooden\_cube} and \texttt{Table}, respectively. As of now, we do not deploy any object recognition algorithm. Therefore the object type must be manually provided when the environment is set up. A list of all the object types and instances that we used can be found in Table~\ref{tab:objects}.

We collected our demonstrations from a virtual reality (VR) environment~\cite{BatesRIC17}.
Using a virtual environment for demonstration allows for easy tracking of the environment, e.g., hand and object positions. However, the downside is a slight loss in realism regarding the haptic experience when interacting with objects. For example, the weight of an object cannot be felt in VR.
We used Unity 
for the creation of the VR environments as well as for simulating the robot task execution and the Valve Index headset to interact in VR. The communication between Unity and all planning/robot-specific code, which runs on several ROS nodes on a Linux machine, is handled through Sigverse~\cite{sigverse}. As our robotic platform, we work with the mobile one-arm manipulator TIAGo.
\begin{table}[h!]
\begin{center}
\caption{Object categories and their instances.}
\small
\begin{tabular}{| l | l | l |}
\hline
  Obj. Types & Instances (learning) & Inst. (execution) \\ 
  \hline
  \hline
 $\textit{Hands}$ & $\{\text{Right\_hand}, \text{Left\_hand}\}$ & $\{ \text{Robot\_gripper}\}$ \\  
 \hline
  & $\{\text{Cube\_red1}, \text{Cube\_green1}$ & $\{\text{Cube\_green3},$  \\
   $\textit{Wooden}$ & $\text{ Cube\_yellow1}, \text{Cube\_blue1},$ & $\text{ Cube\_yellow3},$ \\
  $\textit{\_cubes}$ & $\text{ Cube\_red2}, \text{Cube\_green2},$ & $\text{ Cube\_blue3},$  \\
& $\text{ Cube\_yellow2}, \text{Cube\_blue2}\}$ & $\text{ Cube\_red3} \}$ \\
\hline
 $\textit{Table}$ & $\{ \text{table1} \}$ & $\{ \text{high\_table} \}$ \\
 \hline
 \end{tabular}
\label{tab:objects}
\end{center}
\end{table}

\subsection{Symbolic Planning}
\label{sec:pddl}
Symbolic planning aims at generating a sequence of high-level actions to reach a desired goal. A planning task is split into \texttt{domain} and \texttt{problem}.
The planning \texttt{problem} defines the initial state and the goal. The \texttt{domain} contains a list of all possible actions, also called operators, that can be used in order to reach the goal. Our objective with this paper is to generate the \texttt{domain} automatically from the demonstrations.
There are several ways to formulate a planning task, and PDDL~\cite{pddl} is of the most widespread ones. PDDL was developed in an effort to standardize the formulation of planning tasks. Supported by the popularity of the International Planning Competitions (IPC), a large variety of planners have been and are continuously developed that support PDDL as input. For these reasons, we decided to use PDDL to assess our proposed automatic extraction of operators. Nevertheless, our method can be potentially used with different symbolic planners, e.g., Sequence Planner~\cite{sp}. A \texttt{planning domain} \cite{AP_book} is formally defined as a triple $\Sigma = (S, A, \gamma)$ or a 4-tuple $\Sigma = (S, A, \gamma, cost)$, where $S$ is a finite set of \textit{states}, $A$ is a finite set of actions, $\gamma : S \times A$ a partial function called the \textit{state-transition function} and $cost:S \times A \rightarrow [0, \inf)$ the cost function. 
\subsubsection{State Description}
In a PDDL domain, the world state $S$ is described in terms of a set of state variables that can be either true or false. These state variables describe relations between hands or objects in the environment~\cite{AP_book} like ${\tt inHand}(\texttt{Right\_hand}, \texttt{Cube\_green1})$. In our case, the range over which the state variables can be defined are instances of the object categories that are listed in Table~\ref{tab:objects}.
The full set of state variables that we utilize and the corresponding grounding in terms of continuous input data is listed in Table~\ref{tab:stateVar}. We differentiate between hand specific state variables like \texttt{inHand} or \texttt{handMove} and environment state variables like \texttt{inTouch}. Hand state variables describe either the hand or a relation between the hand and its environment, whereas environment state variables describe relations between objects like cubes and tables. 
\begin{table*}[h!]
\begin{center}
\caption{Shows the defined state variables (\texttt{sv}), and their respective grounding during demonstration and execution. The object categories inside the brackets after the sv-name indicate the range. \texttt{Cube} stands for the category \texttt{Wooden\_Cube} and $\texttt{Thing}=\texttt{Table} \cup \texttt{Wooden\_cube}$.}
\label{tab:stateVar}
\begin{tabular}{ l l l}
\toprule
 State variables (${\tt sv}$) & Grounding & Example (Instantiation) \\ 
 \midrule
  \multicolumn{2}{ l }{\textbf{Hand --} ${\tt sv}$} \\
${\tt inHand}(\texttt{Hand},\texttt{Cube})$ & A hand/gripper has closed its fingers around an object. & ${\tt inHand}(\texttt{Left\_Hand},\texttt{Cube\_red1})$ \\
${\tt actedOn}(\texttt{Hand},\texttt{Cube})$ &
Dist. betw. object and hand $<0.16$m \& hand moving towards obj. & ${\tt actedOn}(\texttt{Right\_Hand},\texttt{Cube\_blue2})$ \\
${\tt handMove}(\texttt{Hand})$ & 
Hand is moving $>0.1 \text{m}/\text{s}$ & ${\tt handMove}(\texttt{Right\_Hand})$ \\
${\tt graspable}(\texttt{Hand},\texttt{Cube})$ & 
Distance between an object and the hand $<0.1$m. & ${\tt graspable}(\texttt{Right\_Hand},\texttt{Cube\_yellow1})$\\
${\tt handOpen}(\texttt{Hand})$ &
Hand is open. & ${\tt handOpen}(\texttt{Left\_Hand})$\\
\midrule
\multicolumn{2}{ l }{\textbf{Environment --} ${\tt sv}$} \\
${\tt inTouch}(\texttt{Thing},\texttt{Thing})$ & 
Unity detects contact between 2 objects. & ${\tt inTouch}(\texttt{Cube\_blue2},\texttt{table1})$ \\
${\tt onTop}(\texttt{Thing},\texttt{Thing})$ & 
Object A on top of object B if inTouch and A higher than B. & ${\tt onTop}(\texttt{Cube\_red2},\texttt{Cube\_green1})$ \\  
\bottomrule
\end{tabular}
\end{center}
\end{table*}

A technicality of our system is that we need to switch between two different state variable representations. While \texttt{handOpen} and \texttt{handMove} are always binary, the rest of the state variables are naturally multi-variate, hence, one of several possible values can be assigned. For example, ${\tt inHand}(\texttt{Right\_hand}) = \texttt{Cube\_green1}$. Some planning description languages like PDDL, however, require a transition into binary state variables.
Concretely that would require the system to map a state variable assignment like ${\tt inHand}(\texttt{Right\_hand}) = \texttt{Cube\_green1}$, which can be expressed with just one single assignment, into a chain of atomic formulae:
\begin{align*}
   {\tt inHand} & (\texttt{Right\_hand}, \texttt{Cube\_green1}) \land \\
    \neg {\tt inHand} & (\texttt{Right\_hand}, \texttt{Cube\_red1}) \land \\
    \neg {\tt inHand} & (\texttt{Right\_hand}, \texttt{Cube\_yellow1}) \land\ \dots 
  \end{align*}
\subsubsection{Planning Operator}
The set of actions $A$ of the planning domain are provided in terms of planning operators $O$.
Operators are blueprints of actions that, if applicable, change the world state in a specific way.  
Each planning operator has an associated \texttt{name}, a set of \texttt{objects} constituting its arguments, a set of \texttt{preconditions} governing what must be true about the world for the operator to be used, and a set of \texttt{effects} describing how the world will change after we use this operator~\cite{ADL19}. \texttt{Preconditions} and \texttt{effects} are expressed in terms of the introduced binary state variables, also called predicates.
An example of the operator with the name \texttt{Operator1} would look as follows:
\begin{align*}
    &\text{Operator1}(\texttt{Hand}, \texttt{Table}, \texttt{Cube}): \\
    &\quad\text{preconditions}:  \quad \quad \quad\quad \quad \quad \text{effects}:\\
        &\;\;\quad{\tt onTop}(\texttt{Cube},  \texttt{Table}) \quad \quad \quad{\tt onTop}(\texttt{Cube}, \texttt{Table}) \\ 
        &\quad\neg {\tt inHand}(\texttt{Hand}, \texttt{Cube}) \quad \quad \quad{\tt inHand}(\texttt{Hand}, \texttt{Cube})
\end{align*}

\noindent The \texttt{Operator1} operator has as input arguments a hand, a table, and a cube. It can only be used when the cube is on top of the table and not in the hand; as an effect, the cube becomes in the hand but is still on top of the table. Using these sets of conditions and effects, a symbolic planner can find a sequence of operators (and associated motor policies $\pi$) that can be applied in a new environment. In our work, we investigate how to automatically identify these preconditions and effects from demonstrations.

\section{System Description}
\addcontentsline{toc}{section}{System}
\subsection{Activity segmentation and classification}
\label{sec:actClass}

To automatically segment and recognize the demonstrated human activities, we used a state-of-the-art learning method that extracts semantic representations from the demonstrations~\cite{ramirez17AIJ}. First, this learning method automatically segments the continuous hand motions based on a minimalistic subset of hand-specific state variables (see Table \ref{tab:objects}), such as \texttt{inHand}, \texttt{actedOn}, \texttt{handMove}. Then, each segmented hand motion will be labeled with a specific symbolic meaning which will represent one of the recognized activities, i.e., \texttt{IdleMotion}, \texttt{Reach}, \texttt{Put}, \texttt{Take}, and \texttt{Stack}. The mapping between the hand state variables and the recognized activities is done by learning a set of general rules using a C4.5 decision tree. Such rules are enhanced with a First-order-Logic reasoning method, and an ontology system \cite{BatesRIC17}. The learned rules are shown in Table~\ref{tab:class}. One of the main advantages of this semantic-based recognition method is its ability to segment and recognize continuous data without training. This means that for our newly analyzed scenario of stacking cubes, we reused the same set of rules from our previous work~\cite{ramirez17AIJ}. Thus no training was performed in this new scenario. Another advantage of this recognition method is that both the activity labels and the obtained rules are human-readable.

\begin{table}[h]
\begin{center}
\caption{\label{tab:class}Hand activity classification rules. T and F stand for true and false respectively, and $\neg$nil means an object as opposed to no-object (nil).}
\begin{tabular}{| c | c | c | c | c | c |}
\hline
 features & \textit{Stack} & \textit{Idle} & \textit{Reach} & \textit{Put} & \textit{Take} \\ 
  \hline
  \hline
 {\tt handMove} & T & T $\lor$ F & T & T & F\\
 \hline
 {\tt actedOn} & $\neg$nil & nil & $\neg$nil &  nil & nil\\
 \hline
 {\tt inHand} & $\neg$nil & nil & nil & $\neg$nil & $\neg$nil\\
  \hline
\end{tabular}
\end{center}
\end{table}
\subsection{Automatic Generation of Operators}
\label{sec:opgen}
Operator generation takes place according to Algorithm~\ref{alg:operator}. As input the algorithm requires a demonstration $D$, in the form of a list of symbolic states $s_t$ at time $t$, subdivided into hand state $s_{t,h}$ and environment state $s_{t,e}$, and the hand activity classification $a_{t,h}$ for each hand $h \in H$ and time point $t$ of the demonstration. The symbolic state representation of a demonstration is generated online based on the grounding rules of Table~\ref{tab:stateVar}. Additionally to the hand/object poses and the hand status, the physical contact information between objects, as detected by the Unity physics engine, is required. The second input is a list of existing operators $O$ obtained from previous demonstrations.

A new operator is generated when the segmentation and classification system detects a transition from one activity to another, $a_{t,h} \neq a_{t-1,h}$ (L-4, Alg. \ref{alg:operator}). For example upon transition from \textit{Reach} to \textit{Take}, the \texttt{Take} operator would be generated, due to changes regarding the ${\tt actedOn}$ and ${\tt inHand}$ state variables. Note that not all state transitions, in particular concerning environment state variables like ${\tt onTop}$, result in a new classification, and consequently not in a new operator since the classification is based on a subset of the state variables.

The advantage of the activity recognition method is that new planning operators can be named automatically with human-understandable labels (L-9, Alg. \ref{alg:operator}), thus providing a semantic description of the under-laying functionality.
The preconditions of an operator are based on the effects of the last state before the activity transition, and effects are based on the last frame of the current activity. For the same example transition from \textit{Reach} to \textit{Take}, preconditions of the \texttt{Take} operator are based on the last state from \textit{Reach} and effects are based on the last state of \textit{Take}. In order to account for any hand-state changes,  which do not lead to a new activity classification, operators are not directly added to $O$ but stored in a buffer list  $O_{\text{Buffer}}$ (L-10, Alg. \ref{alg:operator}) and continuously updated (L-13, Alg. \ref{alg:operator}).

The transition from multi-variate state variables to operator preconditions/effects is carried out in two steps. The first step is the transition from multi-variate state to binary state variables. The major challenge is to exclude any information which is not relevant for a specific operator (L-7, Alg. \ref{alg:operator}). Consider the demonstration domain, which contains eight different cubes. A state variable assignment like ${\tt inHand}(\texttt{Right\_hand}) = \texttt{Cube\_green1}$ implicitly contains the information that no other cube except \texttt{Cube\_green1} is in the hand. We would need seven additional predicates that negate the remaining cubes being in the right hand. Not only have none of these cubes a direct impact on the first one being {\tt inHand}, but the operator would also not be applicable in the execution domain with only three different cubes on the table. Our strategy for choosing the relevant environment predicates only considers the ones that change their value during the operator's application. For the hand state variables, we additionally consider predicates that remain true throughout the activity. The second step is the generalization from the specific cube, e.g., \texttt{Cube\_green1}, to the cube's type, e.g., \texttt{Wooden\_cube} (L-8, Alg. \ref{alg:operator}). This step is based on the assumption that any activity applied to a specific cube, can be performed on every object of the same type.

Currently we only support non-parallel hand demonstrations. Therefore, any environment state variable updates (e.g. ${\tt onTop}$(\texttt{Cube1, Table1}) to $\neg ({\tt onTop}$(\texttt{Cube1, Table1}))), are assigned to the activity performed by the currently active hand (L-15, Alg. \ref{alg:operator}). Consequently, additional preconditions and effects that take this environment change into account are added to the last operator of the responsible hand (L-16, Alg. \ref{alg:operator}).

The last step is updating the operator list $O$ with the newly observed operators from the buffer $O_{\text{Buffer}}$. Each newly observed operator is either added, in case it is different from any other operator (L-19, Alg. \ref{alg:operator}), or the count is incremented for operators who have been already observed before (L-21, Alg. \ref{alg:operator}). 
\setlength{\textfloatsep}{2pt}
\begin{algorithm}
  \caption{\label{alg:operator}Operator generation \& collection}
  \hspace*{\algorithmicindent} \textbf{Input:} demonstration $D=[s_1, a_1, s_2, a_2,... s_n, a_n]$, current list of operators $O$ \\
  \hspace*{\algorithmicindent} \textbf{Output:} updated list of operators $O$
  \begin{algorithmic}[1]
  \State $O_{\text{buffer}} \leftarrow \{\}$
  \For{$t \gets 1, n$} \Comment{$n = |D|$} 
  \ForAll{$h \in H$} 
  \If{$a_{t,h} \neq a_{t-1,h}$} 
  \State $\textit{pre} \gets$ \textsc{StateVarToPredicate}$(s_{t-1,h})$
  \State $\textit{eff} \gets$ \textsc{StateVarToPredicate}$(s_{t,h})$
  \State \textsc{RemoveIrreleventPredicates}$(pre, \textit{eff})$
  \State \textsc{GeneralisePredicates}$(\textit{pre}, \textit{eff})$
  \State $o \gets $ \textsc{Operator}$(a_{t,h}, \textit{pre}, \textit{eff})$ 
  \State \textsc{Append}($O_{\text{buffer}}, o$)
  \Else
  \If{$s_{t,h} \neq s_{t-1,h}$} 
  \State \textsc{UpdateOperator}$(O_{\text{buffer}_h}[-1], s_{t,h})$ 
  \EndIf
  \EndIf
  \EndFor
  \If{$s_{t,e} \neq s_{t-1,e}$} 
  \State $h \gets$ \textsc{ConnectEnvChangeToHand()}
  \State \textsc{AddEnvPredicates}$(O_{\text{buffer}_h}[-1], s_{t,e}, s_{t-1,e})$
  \EndIf
  \EndFor
  \ForAll{$o \in O_{\text{buffer}}$} 
  \If{$o \in O$}
  \State \textsc{IncrementCount}$(O, o)$  
  \Else 
  \State\textsc{Append}$(O, o)$
  \EndIf
  \EndFor
  \end{algorithmic}
\end{algorithm}

\subsection{Plan Quality}
When having a system that continuously learns and collects new skills, inevitably, there will be the possibility to do the same thing in different ways.
How should all the different operators be prioritized?
We propose that this priority should be based on the number of times the operator appeared in the demonstrations.
Our underlying hypothesis is that more commonly observed actions are more frequently used by humans, and second, in the light of potential errors during the classification, also more robust.

Planning problems are often formulated as a minimization of the plan length, with unit operator cost. Some planners like the Fast Downward planner, however, also support cost minimization, with positive integer-valued operator costs \cite{fastdownward}. The conversion from count to cost is done based on the following formula: 
\begin{align*}
    \text{Cost}(op) = \lceil 100(1 - \frac{op_{count}}{op\_type_{count}}) \rceil, 
\end{align*}
where $ op_{count}$ denotes the number of times a specific operator has been observed and $op\_type_{count}$ is the number of times operators of the same type have been observed. As an example, \textit{Stack2} might have been observed 4 times, but overall 20 \textit{Stack} activities were demonstrated, then $\text{Cost}(\textit{Stack2}) = \lceil100(1-4/20)\rceil = 80$.

\subsection{Symbolic Planning and execution}
\label{sec:symbolicplanning}
During run time, the operator information is parsed automatically into a \texttt{problem.pddl} file. Additionally, the \texttt{problem.pddl} file is generated, which requires connection to the execution environment, and a manual goal state definition handled through a ROS service. The initial planning state and object information is then automatically parsed from the current world state. The next step is to find a sequence of actions that will drive the robot from its current state to the goal state. For this reason, a planner is embedded into the execution process. The advantage of having specified the planning problem with PDDL is that a wide variety of planners are available. If we want to contribute from the operator prioritization through cost minimization, the only requirement on the planner is that it supports the \textit{:action-cost} feature from the PDDL3.1 specification. For that reason, we chose the widely-used Fast Downward planning system~\cite{fastdownward}. Upon successful plan generation, that plan is automatically parsed into vector format and prepared for execution.

We assume that a low-level execution function for all operators is available for each operator in the final plan. We execute each function in an open loop, assuming that the execution is always successful. 
For this paper, we have manually implemented the execution functions that automatically retrieve object positions from the execution data flow.

\section{Experiment and results}
\addcontentsline{toc}{section}{Experiment and results}

\subsection{Data collection and operator generation} \label{sec:data_collection}
To test the functionality of the system, three participants were asked to demonstrate how to stack cubes inside our VR environment. Before the first demonstration, each person had the chance to familiarize themselves with the environment. After this introduction phase, four different demonstrations per participant were retrieved, based on the instructions to stack one/two cube(s) with left/right hand. Note that no further requirements were stated, e.g., how fast to perform the activities or which specific cubes to stack.

\begin{figure}[ht!]
    \centering
      \includegraphics[width=0.45\textwidth]{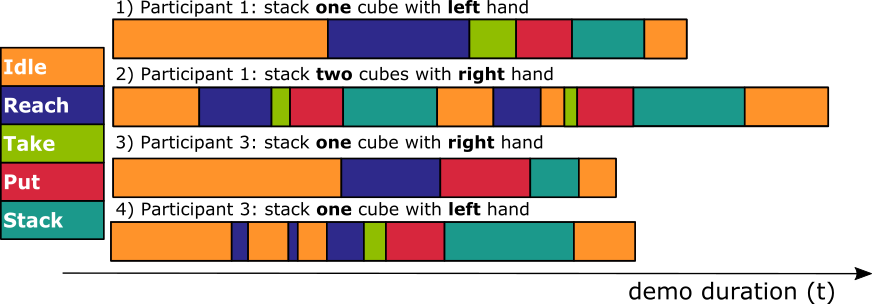}
      \caption{Example demonstrations from 4 different trials of two participants. Each trial is unique in terms of the observed action sequences.}
    \label{fig:demo}
\end{figure}

Figure~\ref{fig:demo} displays four examples of how our system segmented and classified some of the demonstrations. These examples clearly show differences regarding the performed action sequences, which not only occur between different participants but even within different trials of the same participant.

\begin{figure}[ht!]
    \centering
      \includegraphics[width=0.47\textwidth]{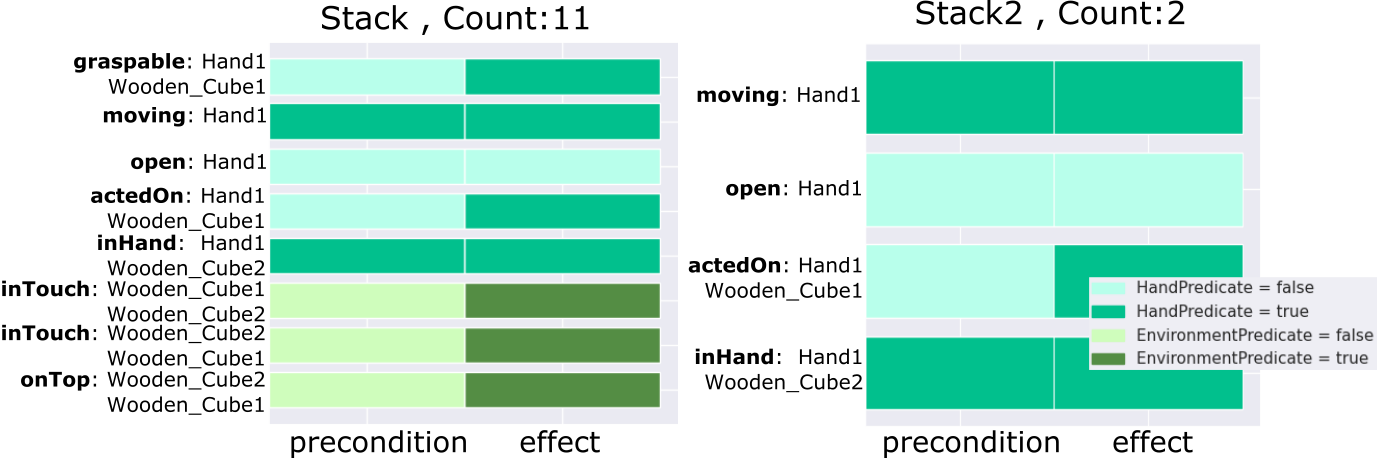}
      \caption{Example of the predicates associated with two different configurations of the \textit{Stack} activities.}
    \label{fig:granType}
\end{figure}

From the 12 demonstrations, in total, 115 operators were automatically extracted (Table~\ref{tab:operatorstatistics}). 
The operator type of each of the detected activities corresponds to the underlying activity classification.
For each operator type, our algorithm found at least four different precondition-effect configurations. We can see that the operator \texttt{IdleMotion} has the largest variety, with 12 different configurations. There are several reasons for this variety. The main reason is that the classification is only based on a subset of the hand state variables (Table~\ref{tab:class}). On top of that, we saw differences regarding demonstration style and speed among the different participants, which is reflected in their demonstration segmentation patterns. Consider, for example, that the activity \textit{Take} is not part of the third demonstration of Fig.~\ref{fig:demo}. The overall length of this specific demonstration is shorter, indicating a higher hand velocity, and the cube was taken in one fluent hand movement without stopping, which would have been a requirement for a \texttt{Take} classification (Tab.~\ref{tab:class}). Nevertheless, our system is able to successfully handle this demonstration by including the update of the \texttt{inHand} state variable into the \texttt{Put} operator.
Thus we observed some \texttt{Put} operators, which include a change in the \texttt{inHand} predicate, and some without.
In general, our segmentation and recognition accuracy lies above 80 percent \cite{ramirez17AIJ}. Particularly challenging is finding the exact end and start timings during activity transitions. This could lead to varying allocations of environment changes in cases where they happen during these transition phases. An example for different \texttt{Stack} operator configurations is shown in Fig.~\ref{fig:granType}.

\begin{table}[h]
\begin{center}
\caption{\label{tab:operatorstatistics} Operator statistics: for each operator type, different precondition/effect configurations were observed.}
\begin{tabular}{| l | c | c |}
\hline
 operator type & \# op. configurations & operator count \\ 
 & (per type)  & (per type) \\
 \hline
 \hline
 Stack  & 4  & 16\\
 IdleMotion  & 12  & 31 \\
 Put  &  5 & 22\\
 Reach  & 5  & 30\\
 Take  &  4 &  16\\
 \hline
 total count  & 30  & 115\\
 \hline
\end{tabular}
\end{center}
\end{table}

\subsection{Plan generation}
\begin{figure}[ht!]
    \centering
      \includegraphics[width=0.45\textwidth]{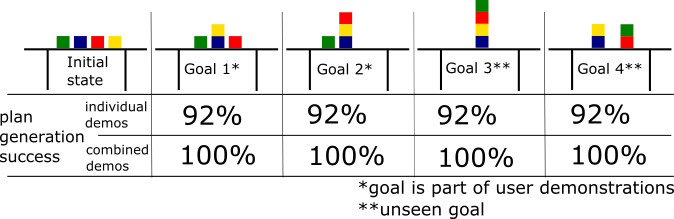}
      \caption{Plan goals with corresponding plan generation success ratio.}
    \label{fig:plangoals}
\end{figure}

Four different stacking goals are utilized to evaluate the operator generation process (Fig. \ref{fig:plangoals}).
The first two goals have been part of the experiment instructions, albeit the participants were free in their cube choice. Goal 3 and Goal 4, however, are two new cube configurations. For all the goals, plans are generated by calling the Fast-Downward planner, based on the extracted operators. We performed two experiments: I) we generate plans using domains from only one demonstration at a time (i.e., individual demos), and II) we generate plans using the operators obtained from a domain that incorporates all 12 demonstrations together (i.e., combined demos). We observed that if we only use the information from one demonstration, then the plan generation was successful for 11 out of the 12 individual demonstrations (92\%, Fig. \ref{fig:plangoals},  second row).
Furthermore, we observed that for using the operators obtained from all the demonstrations combined, plan generation was successful in all cases (Fig. \ref{fig:plangoals}, third row).

\begin{figure*}[]
    \centering
      \includegraphics[width=0.95\textwidth]{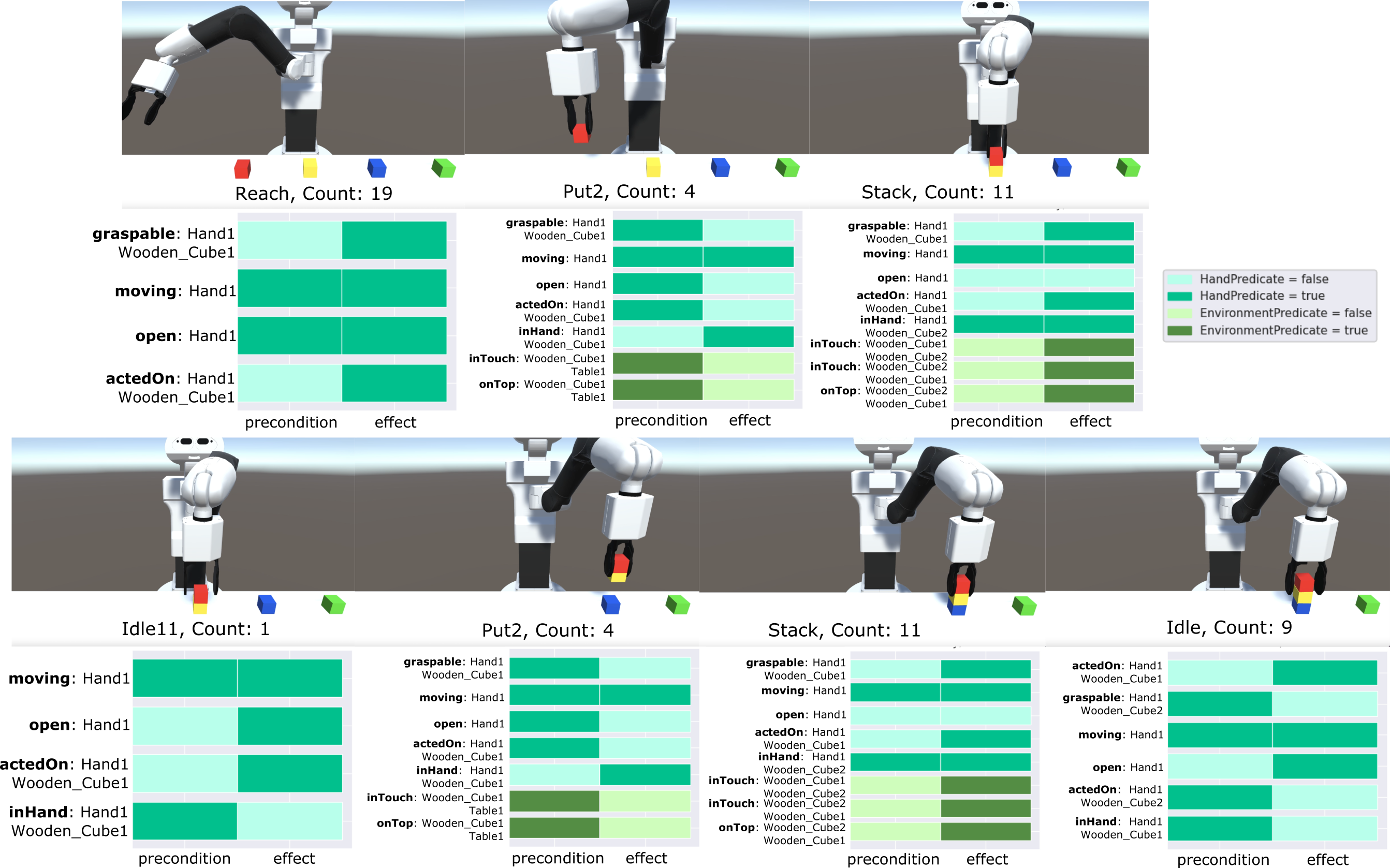}
      \caption{Plan Execution of a simple stacking goal.}
    \label{fig:planexec}
\end{figure*}

These results show that in most cases, one single demonstration is sufficient to generate plans to satisfy the goals, but more data contributes to a higher success probability. The complete plan for Goal 2 is illustrated in Fig. \ref{fig:planexec}. To observe the execution in motion, please refer to our video\footnote{https://youtu.be/hEUEpQcrDtw}.

It is also worthwhile to mention that plans are not just repetitions of the observed sequence from the demonstration. For example, we can observe from Fig. \ref{fig:demo}, that the bottom demonstration shows two initial \textit{Reach-Idle} sequences, which have no meaningful contribution to the goal satisfaction. 
Nevertheless, these useless sequences where avoided by the planner, and the action sequence of \texttt{Reach}, \texttt{Take}, \texttt{Put}, \texttt{Stack}, \texttt{Idle} was generated.

The plan cost is evaluated for the combined-demos domain. Plans for Goals $1-4$ are calculated with and without cost optimization. The plan cost was reduced by approximately 9\%,  9\%, 16\%, and 13\%, respectively, which accounts for a plan improvement of 11.75\% averaged over all four evaluated plans.

\subsection{Plan execution}
While plan generation is successful in most cases, not all plans are guaranteed to be executable. The bottom line is that generalization is no free lunch, and occasionally the planner exploits consequent loopholes that come up at the transition from multi-variate state variables to binary predicates (L-7, Algo. \ref{alg:operator}). This can be illustrated on the \texttt{Reach} operator (first step of the plan execution in Fig. \ref{fig:planexec}). Application of this operator would result for example in \texttt{Cube\_green1} being ${\tt actedOn}$ and ${\tt graspable}$ by \texttt{Hand\_right}. Applying two reach activities in a row is theoretically a perfectly fine action sequence, and even though no observation of this exact sequence was made, the planner occasionally suggested this move in several of the generated example plans.
Going back to the example, if, after reaching for \texttt{Cube\_green1},  \texttt{Reach} would be applied on \texttt{Cube\_yellow1}, both cubes are ${\tt actedOn}$ and ${\tt graspable}$ as a result. This is, in reality, not true, of course, but negating any potential previous ${\tt actedOns}$ and ${\tt graspables}$ are not captured by the operator. 
The planner suggested then to apply the \texttt{Take} operator on \texttt{Cube\_green1}, which is in reality not  ${\tt graspable}$ anymore, so the plan execution failed.

This problem can be easily fixed in an automated manner with the additional domain knowledge that only one object is graspable at any given time. 
We add the negation of all previous ${\tt actedOn}$s and ${\tt graspable}$s as an effect to all operators which have a transition from $\neg {\tt actedOn}$(\texttt{Hand1}, \texttt{Cube1}) to ${\tt actedOn}$(\texttt{Hand1}, \texttt{Cube1}), or from $\neg {\tt graspable}$(\texttt{Hand1}, \texttt{Cube1}) to ${\tt graspable}$(\texttt{Hand1}, \texttt{Cube1}).
All eight previously non-executable plans (out of the 60 individual-demo based plans) could be fixed in this way.

The following is an excerpt from the automatically generated PDDL domain file:
\begin{lstlisting}[
    basicstyle=\small, %or \tiny or \footnotesize etc.
]
(define (domain 
    learningFromDemonstrationAllOperators)
(:requirements :strips :typing 
    :negative-preconditions :action-costs)
(:types Wooden_cube - Thing Hand - Thing 
        Table - Thing)
(:predicates
   (inHand ?Hand1 - Hand 
           ?Wooden_cube1 - Wooden_cube)
   (actedOn ?Hand1 - Hand 
                ?Wooden_cube1 - Wooden_cube)
   (handOpen ?Hand1 - Hand)
   (handMove ?Hand1 - Hand)
   (onTop ?Thing1 - Thing ?Thing2 - Thing)
   (inTouch ?Thing1 - Thing ?Thing2 - Thing)
   (graspable ?Hand1 - Hand ?Thing1 - Thing)
  )

(:functions (total-cost))
		
(:action Stack
  :parameters (?Hand1 - Hand  
               ?Wooden_cube1 - Wooden_cube  
               ?Wooden_cube2 - Wooden_cube)
  :precondition (and 
   (not(inTouch ?Wooden_cube1 ?Wooden_cube2)) 
   (not(inTouch ?Wooden_cube2 ?Wooden_cube1)) 
   (not(onTop ?Wooden_cube2 ?Wooden_cube1)) 
   (inHand ?Hand1 ?Wooden_cube2)
   (not(actedOn ?Hand1 ?Wooden_cube1)) 
   (not(handOpen ?Hand1)) 
   (handMove ?Hand1) 
   (not(graspable ?Hand1 ?Wooden_cube1)) 
   (not(= ?Wooden_cube1 ?Wooden_cube2)) 
   (not(= ?Wooden_cube2 ?Wooden_cube1)))
  :effect (and 
   (inTouch ?Wooden_cube1 ?Wooden_cube2) 
   (inTouch ?Wooden_cube2 ?Wooden_cube1) 
   (onTop ?Wooden_cube2 ?Wooden_cube1) 
   (inHand ?Hand1 ?Wooden_cube2) 
   (actedOn ?Hand1 ?Wooden_cube1) 
   (not(handOpen ?Hand1)) 
   (handMove ?Hand1) 
   (graspable ?Hand1 ?Wooden_cube1) 
   (increase (total-cost) 31))
 )
...
(:action Put
  :parameters (?Hand1 - Hand  
               ?Table1 - Table  
               ?Wooden_cube1 - Wooden_cube )
  :precondition (and 
   (inTouch ?Wooden_cube1 ?Table1) 
   (onTop ?Wooden_cube1 ?Table1) 
   (inHand ?Hand1 ?Wooden_cube1) 
   (not(handOpen ?Hand1)) 
   (not(handMove ?Hand1)))
  :effect (and 
   (not(inTouch ?Wooden_cube1 ?Table1)) 
   (not(onTop ?Wooden_cube1 ?Table1)) 
   (inHand ?Hand1 ?Wooden_cube1) 
   (not(handOpen ?Hand1)) 
   (handMove ?Hand1) 
   (increase (total-cost) 45))
 )
...
)
\end{lstlisting}

\subsection{Discussion}
As a next step, we want to scale our system to more complex tasks. One of the biggest bottlenecks of our method, and symbolic planning in general, is the state variable selection. Currently, we follow the traditional method~\cite{Agostini2020} of handpicking a set of task-specific variables and manually defining their grounding functions. While it is an advantage that our hand activity recognition only relies on a small set of predicates, we will likely need to grow the set of environment predicates when we want to describe more complex interactions with the environment. One strategy could be to commit to an as-general-as-possible base set of state variables, which could comprise the set of object-centered spatial state variables proposed in~\cite{Agostini2020}. This would, however, still required a manual addition of domain-specific features. Another option would be to learn the state variable choice as well as their grounding as it is done in~\cite{Ahmetoglu2020}.
\section{Conclusion}
\addcontentsline{toc}{section}{Discussion and conclusion}
In this paper, we presented a system for automated planning domain generation, which allows robots to accomplish new tasks based on human demonstrations. Our system analyzes the transitions of the classified activities and extracts the relevant preconditions and effects. As much as one single stacking demonstration can be enough (92\%) to allow for successful plan generation of various unseen stacking scenarios. The system also allows for continuous skill collection and prioritizes more often observed operators based on the operator cost minimization during the planning process.

\addtolength{\textheight}{-12cm}   




\section*{Acknowledgment}
\addcontentsline{toc}{section}{Acknowledgment}
The research reported in this paper has been supported by Chalmers AI Research Centre (CHAIR).

\bibliography{mybib}
\bibliographystyle{IEEEtran}

\end{document}